\documentclass[runningheads]{llncs}

\usepackage[T1]{fontenc}
\usepackage{graphicx}
\usepackage{amsmath}
\usepackage{amssymb}
\usepackage{booktabs}
\usepackage{multirow}
\usepackage{cleveref}
\usepackage{algorithm}
\usepackage{algpseudocode}
\usepackage{setspace}
\usepackage{enumitem}
\usepackage{microtype}
\usepackage{subcaption}
\usepackage{marvosym}
\usepackage{placeins}  % 提供 \FloatBarrier 命令
\usepackage{stfloats}  % 改善底部浮动体处理(可选)
\usepackage{array}
\newcommand{\thickhline}{\noalign{\hrule height 1pt}}
\begin{document}
\title{BCP: A Lightweight Boundary-Controlled Framework for Online Sign Language Recognition and Translation}
\titlerunning{Lightweight Boundary-Controlled Framework for Online SLR and SLT}
% If the paper title is too long for the running head, you can set
% an abbreviated paper title here
%
\author{Xiao Ma\inst{1} \and
Lin Li\inst{1}\textsuperscript{(\Letter)} \and
Shiqing Wu\inst{2} \and
Wang Gao\inst{3} \and
Chuanbo Liu\inst{1} \and
Jianquan Liu\inst{4}}
\authorrunning{X. Ma et al.}
% First names are abbreviated in the running head.
% If there are more than two authors, 'et al.' is used.
%
\institute{
Wuhan University of Technology, Wuhan, China\\
\email{361314@whut.edu.cn, cathylilin@whut.edu.cn, Lchb72@whut.edu.cn}
\and
City University of Macau, Macau, China\\
\email{sqwu@cityu.edu.mo}
\and
Jianghan University, Wuhan, China\\
\email{gaow@jhun.edu.cn}
\and
Biometrics Research Laboratories, NEC Corporation, Japan\\
\email{jqliu@nec.com}
}

\maketitle              % typeset the header of the contribution
\begin{abstract}
Online sign language recognition (SLR) and translation (SLT) incrementally produce gloss sequences or natural language text from sign video as it arrives, a capability essential for real-time communication and live captioning. 
However, powerful pretrained offline models cannot be directly deployed in this setting, while existing online methods typically retrain the entire model on simulated streaming prefixes constructed from offline videos, incurring high costs. 
To address this challenge, we propose \textbf{\underline{B}oundary-\underline{C}ontrolled \underline{P}refix Commit} (BCP), a unified lightweight framework that converts pretrained offline methods into online counterparts for sign-to-gloss (S2G), sign-to-text (S2T), and sign-to-gloss-to-text (S2G2T). 
Specifically, BCP freezes the pretrained model and introduces a lightweight boundary adapter that decides when to commit, supervised by Connectionist Temporal Classification (CTC) alignments from a pretrained SLR model, without any human annotation. At inference, a Boundary-Controlled Wait-K Output Policy then decides what to commit via prefix voting, yielding adaptive burst commits while preserving the output progress of the base Wait-K schedule. 
Experiments on PHOENIX-2014T and CSL-Daily across four offline methods show that BCP, with less than 0.1\% additional trainable parameters per offline method, achieves a strong quality--latency trade-off on S2T and remains competitive on S2G and S2G2T.
\end{abstract}

\section{Introduction}
\label{sec:intro}

\begin{figure}[t!]
    \centering
    \includegraphics[width=1.0\textwidth]{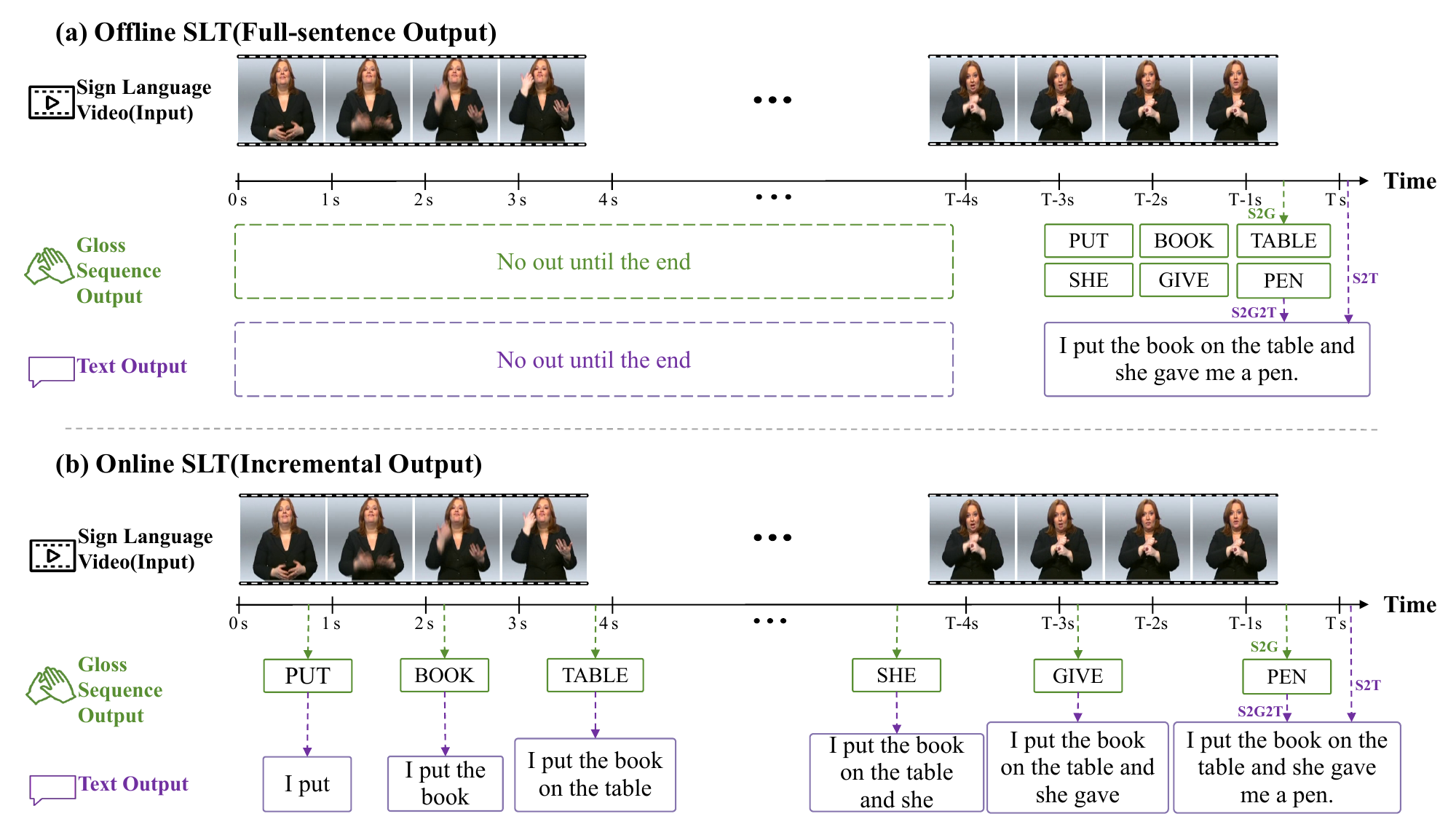}
    \caption{Comparison between offline and online sign language tasks (S2G, S2T, S2G2T). (a) Offline tasks generate outputs only after the full video ends. (b) Online tasks incrementally produce outputs as the video unfolds.}
    \label{fig:intro}
\end{figure}

Sign language recognition and translation bridge deaf and hearing communities, with applications in assisted communication, education, and live captioning~\cite{camgoz2018neural,camgoz2020slt}. In these scenarios, translation must keep pace with sign production rather than wait for the entire utterance, motivating online sign language recognition and translation that incrementally produce gloss sequences or text as the video unfolds (Figure~\ref{fig:intro}(b)), in contrast to offline counterparts that decode only after the full video is observed (Figure~\ref{fig:intro}(a)).

Recent offline methods built on strong visual encoders~\cite{carreira2017i3d}, keypoint modeling~\cite{chen2022twostream}, and pretrained translation models~\cite{chen2022mmtlb,liu2020mbart,zhao2024cvslt} have substantially improved performance, yet they rely on bidirectional context and full-sequence attention and thus cannot be directly deployed for online use. Existing online methods~\cite{sun2024ctl,yin2021simulslt,zuo2024online} address this limitation by retraining dedicated online models on simulated streaming prefixes. This strategy requires task-specific data construction and full-model retraining, resulting in substantial cost and limited reuse of the rapidly evolving offline ecosystem.

Our key insight is that adapting an offline sign language system to online use does not require relearning recognition or translation capability. It is effective to apply lightweight control to two coupled decisions: when to commit and what to commit under partial observations. Based on this insight, we propose BCP, a unified lightweight framework that converts pretrained offline SLR, SLT, and gloss-to-text models into online counterparts for S2G, S2T, and S2G2T tasks. Specifically, Boundary-Controlled specifies when to commit: each commit is triggered by a predicted sign-segment boundary. Prefix Commit specifies what to commit: the longest stabilized prefix across consecutive predictions is committed. BCP freezes all pretrained backbones and trains only a small boundary adapter, with no human annotation required.

Our BCP consists of two modules. First, a boundary adapter learns to predict sign-segment boundaries on top of the frozen SLR encoder, supervised by CTC alignments from a pretrained SLR model. Second, we introduce a Boundary-Controlled Wait-K Output Policy that combines a boundary-controlled Wait-K fallback schedule with longest common prefix (LCP) voting. LCP voting commits content once it becomes stable across consecutive decoding steps, while the Wait-K fallback commits the next token when no additional stabilized prefix is available. This policy introduces no additional trainable parameters.

On PHOENIX-2014T and CSL-Daily across four offline methods, BCP achieves a strong quality--latency trade-off on S2T and remains competitive on S2G and S2G2T, with only 0.13M additional trainable parameters (under 0.1\% of each offline method).

Our main contributions are:
\begin{itemize}[leftmargin=*]
    \item We formulate online sign language generation as two coupled decisions, when and what to commit under partial video input, and propose BCP, a lightweight framework converting pretrained offline models into online counterparts for S2G, S2T, and S2G2T without retraining the offline backbones.
    \item We design a boundary adapter on a frozen SLR encoder to predict sign-segment boundaries. Supervised by CTC alignments from a pretrained SLR model without human annotation, it provides timing signals for online commitment.
    \item Experiments on PHOENIX-2014T and CSL-Daily across four offline methods show that BCP achieves a strong quality--latency trade-off against prior online baselines while keeping the offline backbones frozen.
\end{itemize}

\section{Related Work}
\label{sec:related}
\subsection{Sign Language Recognition and Translation}
SLR and SLT research focuses on offline modeling, formulating SLT as sequence-to-sequence neural machine translation~\cite{camgoz2018neural,camgoz2020slt,vaswani2017attention}. Advances in visual encoders, keypoint representations, and pretrained language models, such as MMTLB~\cite{chen2022mmtlb}, TwoStream-SLR/SLT~\cite{chen2022twostream}, SLTUNet~\cite{zhang2023sltunet}, CV-SLT~\cite{zhao2024cvslt}, STMC~\cite{zhou2022stmc}, and gloss-free variants~\cite{zhou2023gfslt}, yield strong offline performance but remain inherently offline, precluding direct online use.
Online methods address this limitation at significant cost. SimulSLT~\cite{yin2021simulslt} implicitly models gloss boundaries via an integrate-and-fire mechanism with Wait-K. Online SLR methods~\cite{cheng2020fcn,min2021vac,zuo2024online} build sliding-window recognizers and cascade Wait-K gloss-to-text translators. CTL and CTL++~\cite{sun2024ctl} control output by estimating the confidently translatable length. All require constructing simulated streaming prefixes per task and retraining or fine-tuning end-to-end, preventing reuse of the evolving offline ecosystem. In contrast, BCP converts a pretrained offline method into an online counterpart by decomposing adaptation into when-to-commit and what-to-commit decisions with a single lightweight adapter.
\subsection{Simultaneous Translation and Online Control}
BCP recombines three lines of prior work under a unified perspective: Wait-K, CTC forced alignment, and the longest common prefix (LCP) criterion. Wait-K~\cite{ma2019stacl} is a prefix-to-prefix strategy guaranteeing a fixed lag upper bound when used alone, with adaptive variants~\cite{arivazhagan2019milk,ma2020mma,zhang2022gma} reducing effective latency; both assume discrete textual advance units inapplicable to continuous sign videos, which BCP addresses by predicting sign-segment boundaries as advance units. CTC forced alignment~\cite{graves2006ctc} extracts frame-to-symbol correspondences without human annotation, applied to offline gloss localization but not online supervision; BCP repurposes it as boundary pseudo-supervision for an adapter deciding when to commit. The LCP criterion~\cite{sun2024ctl} identifies stabilized content across consecutive decoding steps but lacks explicit progress control; BCP integrates LCP voting with a boundary-controlled Wait-K fallback schedule, so that LCP enables burst commits while Wait-K ensures that BCP never falls behind the base schedule. BCP's contribution lies not in any single primitive, but in the when-to-commit and what-to-commit decomposition that unifies them.

\begin{figure}[t]
    \centering
    \includegraphics[width=1.0\textwidth]{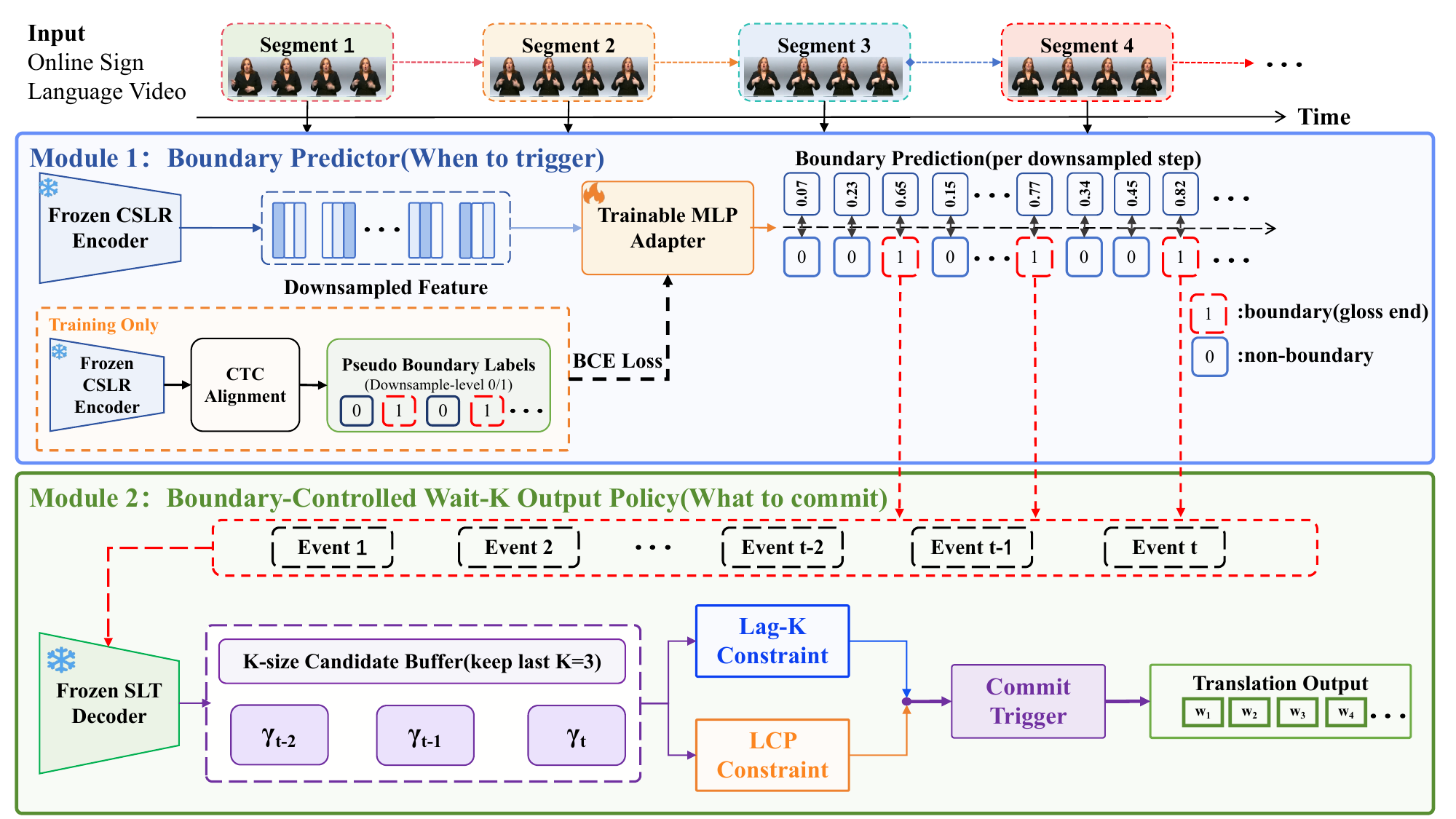}
    \caption{Overall framework of BCP for online sign language recognition and translation. The framework consists of a boundary prediction module and a boundary-controlled Wait-K output policy incorporating LCP-based prefix voting.}
    \label{fig:framework}
\end{figure}

\section{Methodology}

\subsection{Problem Formulation and Framework Overview}
Given an online sign language video $V = \{v_1, v_2, \ldots\}$ whose frames arrive in temporal order, we assume access to frozen offline backbones: a SLR encoder $E_{slr}$ with decoding head $D_{slr}$, an SLT decoder $D_{slt}$ for S2T, and a gloss-to-text model $D_{g2t}$ for S2G2T. The goal of BCP is to produce an online output sequence $O = \{o_1, o_2, \ldots\}$, where each $o_j$ is committed at a certain moment, with a unified form across S2G, S2T, and S2G2T.

The overall architecture of BCP is illustrated in Figure~\ref{fig:framework}. Two modules handle the when-to-commit and what-to-commit decisions: a boundary prediction module and a Boundary-Controlled Wait-K Output Policy. The former, built on the frozen SLR encoder $E_{slr}$, predicts the ending position of each gloss along the downsampled temporal dimension. The latter uses a boundary-controlled Wait-K fallback schedule and incorporates longest common prefix (LCP) voting to decide what to commit at each step. Only the boundary adapter $f_{\theta}$ is trainable.

\subsection{Boundary Adapter with Pseudo Supervision}
The boundary adapter addresses when to commit. We use sign-segment boundaries, i.e., the ending positions of glosses, as commit triggers. Gloss endings provide natural semantic commit points and help avoid breaking partial signs, as also adopted in prior online sign language works~\cite{yin2021simulslt,zuo2024online}. This boundary choice is consistent across our three tasks: glosses are the recognition target in S2G, the alignment anchor in S2T, and the explicit intermediate representation in S2G2T.

Given an online video prefix $V_{1:t}$, the frozen SLR encoder outputs a downsampled feature sequence $H_{1:T} = E_{slr}(V_{1:t}) \in \mathbb{R}^{T \times d}$, where $T = \lfloor t / r \rfloor$ and $r$ is the downsampling rate. Following the spirit of parameter-efficient adaptation~\cite{houlsby2019adapter,hu2022lora}, a lightweight MLP adapter $f_{\theta}$ maps each feature to a boundary probability:
\begin{equation}
    b_{\tau} = \sigma\bigl(f_{\theta}(h_{\tau})\bigr), \quad \tau = 1, \ldots, T
    \label{eq:adapter}
\end{equation}
where $\sigma(\cdot)$ is the sigmoid function and $b_{\tau} \in [0, 1]$ denotes the probability that step $\tau$ is a sign-segment boundary. 

A core challenge is that datasets such as PHOENIX-2014T~\cite{camgoz2018neural} and CSL-Daily~\cite{zhou2021signbt} only provide sentence-level gloss sequences $G = \{g_{1}^{*}, \ldots, g_{N}^{*}\}$ without frame-level labels. To obtain supervision without human annotation, we introduce an auxiliary frozen SLR model $E_{slr}^{aux}$, independent of $E_{slr}$ and possibly differing in architecture or training corpus. Given its visual features and $G$, its decoding head performs CTC forced alignment to obtain the optimal path:
\begin{equation}
    \pi^{*} = \arg\max_{\pi \in \mathcal{A}(G)} \prod_{\tau=1}^{T} p_{aux}(\pi_{\tau} \mid \tau)
    \label{eq:ctc-alignment}
\end{equation}
where $\mathcal{A}(G)$ denotes all CTC paths consistent with $G$ and $p_{aux}$ is the per-step output distribution of $E_{slr}^{aux}$. The pseudo boundary label at step $\tau$ is then:
\begin{equation}
    \hat{b}_{\tau} =
    \begin{cases}
        1, & \text{if } \pi_{\tau}^{*} \in \{g_{1}^{*}, \ldots, g_{N}^{*}\} \text{ and } \pi_{\tau+1}^{*} \neq \pi_{\tau}^{*} \\
        0, & \text{otherwise}
    \end{cases}
    \label{eq:pseudo-label}
\end{equation}
marking $\tau$ as a sign-segment boundary when its aligned gloss differs from the next step. The adapter is trained with class-weighted binary cross-entropy (BCE):
\begin{equation}
    \mathcal{L}_{boundary} = -\frac{1}{T} \sum_{\tau=1}^{T} \left[ \hat{b}_{\tau} \log b_{\tau} + (1 - \hat{b}_{\tau}) \log(1 - b_{\tau}) \right]
    \label{eq:bce-loss}
\end{equation}
where the class weighting alleviates the imbalance from boundary sparsity.

The supervision and the adapter come from two independent pretrained SLR models: $E_{slr}$ provides visual representations for the adapter, while $E_{slr}^{aux}$ produces pseudo labels via CTC forced alignment. This decoupling enables learning sign-segment boundaries without human annotation, and lets the adapter be directly reused across S2G, S2T, and S2G2T.
\subsection{Boundary-Controlled Wait-K Output Policy}
This policy determines what to commit at each boundary event. We use Wait-K as a base progress schedule and add LCP voting to burst-commit stabilized prefixes. We use the boundary event index $i$ as the primary subscript: $i$ denotes the $i$-th detected boundary event, $g_{i}$ the cumulative boundary count, $O_{i}$ the committed sequence, $C_{i}$ the candidate decoding sequence, and $P_{i}^{lcp}$ the longest common prefix over the candidate buffer. A single hyperparameter $K$ controls both the Wait-K waiting length and the LCP voting window size. In validation, the two roles preferred the same scale, so a shared $K$ gives an interpretable one-dimensional control knob without sacrificing performance.

We first adapt Wait-K by replacing its source-side advance unit. The conventional advance unit of Wait-K is a discrete text token, which does not apply to continuous video. BCP redefines the advance unit as a sign-segment boundary detected by the adapter. An event is triggered when $e_{\tau} = \mathbb{I}(b_{\tau} > \tau_{thr})$, and the cumulative count at the $i$-th event is $g_{i} = \sum_{\tau=1}^{T_{i}} e_{\tau}$, where $T_{i}$ denotes the corresponding downsampled step. With this redefinition, $K$ specifies the number of boundary events required before any commit. At the $i$-th event, given $g_{i}$ and the committed token count $|O_{i}|$, Wait-K commits the next single token whenever:
\begin{equation}
    g_{i} \geq K \quad \text{and} \quad |O_{i}| < g_{i} - K + 1
    \label{eq:waitk-condition}
\end{equation}
Let $\ell_i = g_i - K + 1$ be the base Wait-K progress target. In BCP, $\ell_i$ is used as a fallback target rather than a hard limit: after the initial Wait-K gate, the policy first commits stabilized LCP tokens and falls back to one Wait-K token only when no longer stabilized prefix exists. This applies uniformly across S2G, S2T, and S2G2T: even when the target is natural language, the source-side advance unit remains the sign-segment boundary.

\begin{table}[!t]
\centering
\caption{Trainable parameter comparison. ${\sim}$ denotes values estimated from the original papers' reported architectures when exact parameter counts were not provided.}
\label{tab:params}
\footnotesize
\renewcommand{\arraystretch}{1.1}
\setlength{\tabcolsep}{6pt}
\begin{tabular}{l|c|c|c}
\hline
\textbf{Method} & \textbf{S2T} & \textbf{S2G} & \textbf{S2G2T} \\
\thickhline
\multicolumn{4}{l}{\textbf{Online methods (trainable parameters)}} \\
\hline
SimulSLT~\cite{yin2021simulslt} & ${\sim}$40M & - & - \\
\hline
CTL/CTL++~\cite{sun2024ctl} & 29.75M & - & - \\
\hline
SLRT~\cite{zuo2024online} & - & 37.97M & 366.10M \\
\hline
\textbf{BCP (ours)} & \textbf{0.13M} & \textbf{0.13M} & \textbf{0.13M} \\
\hline
\multicolumn{4}{l}{\textbf{Offline backbones (reference, frozen by BCP)}} \\
\hline
MMTLB~\cite{chen2022mmtlb} & 377.52M & 14.59M & 366.10M \\
\hline
TS-SLT~\cite{chen2022twostream} & 436.12M & 122.83M & 366.10M \\
\hline
CV-SLT~\cite{zhao2024cvslt} & 363M & - & - \\
\hline
\end{tabular}
\end{table}

Wait-K alone commits at most one token per boundary event, which can be unnecessarily conservative when the decoder has already produced stable outputs across recent candidates. We introduce a prefix voting mechanism that detects such stability and commits multiple tokens at once, while retaining Wait-K as a fallback when no additional stable prefix is found. At each event, the frozen decoder $D$ produces a candidate $C_{i} = D(\text{visible prefix at the } i\text{-th boundary})$, where $D$ is $D_{slr}$ for S2G, $D_{slt}$ for S2T, and $D_{g2t} \circ D_{slr}$ for S2G2T. The system maintains a sliding buffer $\mathcal{B}_{i} = \{C_{i-K+1}, \ldots, C_{i}\}$ for $i \geq K$ and computes:
\begin{equation}
    P_{i}^{lcp} = \mathrm{LCP}(C_{i-K+1}, \ldots, C_{i})
    \label{eq:lcp}
\end{equation}
where $\mathrm{LCP}(\cdot)$ returns the longest token-wise identical prefix. Intuitively, $P_{i}^{lcp}$ is the content that has stabilized across the last $K$ candidates and is therefore safe to commit. For instance, with $K=3$ and $C_{i-2} = w_{1} w_{2} w_{3} w_{4} w_{5}$, $C_{i-1} = w_{1} w_{2} w_{3} w_{4} w_{6}$, $C_{i} = w_{1} w_{2} w_{3} w_{4} w_{7}$, the three agree on the first four tokens, yielding $P_{i}^{lcp} = w_{1} w_{2} w_{3} w_{4}$.

The two mechanisms are combined into a single commit rule. At each event, the system first checks the minimum Wait-K gate. If $g_i<K$, no token is committed. Otherwise, it compares $P_{i}^{lcp}$ with $O_{i}$: if the stabilized prefix extends the committed sequence, all extra stabilized tokens in $P_{i}^{lcp}$ are committed at once; otherwise, the policy falls back to Wait-K whenever $|O_i|<\ell_i$. The commit content is:
\begin{equation}
    o_{i}^{new} =
    \begin{cases}
        \varnothing, & \text{if } g_i < K \\
        P_{i}^{lcp}[\,|O_{i}|+1 : |P_{i}^{lcp}|\,], & \text{if } O_i \preceq P_i^{lcp} \text{ and } |P_{i}^{lcp}| > |O_{i}| \\
        C_{i}[\,|O_{i}|+1\,], & \text{if } |O_i| < \ell_i \\
        \varnothing, & \text{otherwise}
\end{cases}
    \label{eq:commit}
\end{equation}
After each commit, the committed sequence is updated as $O_{i+1} = O_{i} \oplus o_{i}^{new}$, where $\oplus$ denotes concatenation, and fed back to the frozen decoder. Here $O_i \preceq P_i^{lcp}$ means that $O_i$ is a prefix of the stabilized prefix. Thus, under the same boundary events, BCP never outputs fewer tokens than base Wait-K, while LCP bursts can move beyond $\ell_i$ to reduce effective latency.

\section{Experiments}

\begin{table}[t]
\centering
\caption{S2T results on PHOENIX-2014T and CSL-Daily. R, B4, and AL denote ROUGE-L, BLEU-4, and Average Lagging (ms), respectively. Best online results are shown in bold.}
\label{tab:s2t}
\renewcommand{\arraystretch}{1.1}
\resizebox{\textwidth}{!}{%
\begin{tabular}{l|c|c|c|c|c|c|c|c|c|c|c|c}
\hline
\multirow{3}{*}{\textbf{Method}} & \multicolumn{6}{c|}{\textbf{PHOENIX-2014T}} & \multicolumn{6}{c}{\textbf{CSL-Daily}} \\
\cline{2-7}\cline{8-13}
& \multicolumn{3}{c|}{Dev} & \multicolumn{3}{c|}{Test} & \multicolumn{3}{c|}{Dev} & \multicolumn{3}{c}{Test} \\
\cline{2-13}
& R & B4 & AL & R & B4 & AL & R & B4 & AL & R & B4 & AL \\
\thickhline
\multicolumn{13}{l}{\textbf{Offline methods}} \\
\hline
MMTLB             & 52.92 & 27.37 & -    & 53.12 & 28.61 & -    & 53.20 & 24.20 & -    & 53.31 & 24.05 & -    \\
\hline
TS-SLT(joint)     & 53.02 & 28.08 & -    & 52.72 & 28.23 & -    & 54.01 & 24.59 & -    & 54.46 & 24.65 & -    \\
\hline
TS-SLT(ensemble)  & 54.11 & 28.65 & -    & 53.52 & 28.92 & -    & 55.06 & 25.68 & -    & 55.67 & 25.64 & -    \\
\hline
CV-SLT            & 55.03 & 29.12 & -    & 54.45 & 29.31 & -    & 56.42 & 28.26 & -    & 57.16 & 28.76 & -    \\
\hline
\multicolumn{13}{l}{\textbf{Online methods}} \\
\hline
SimulSLT            & -     & -     & -    & 45.62 & 20.10 & 1905 & -     & -     & -    & 42.63 & 13.44 & 2054 \\
\hline
CTL                 & 48.75 & 22.08 & 1641 & 47.78 & 21.71 & 1647 & 41.12 & 13.26 & 1550 & 40.56 & 13.12 & 1556 \\
\hline
CTL++               & 48.71 & 22.71 & 1684 & 47.72 & 22.67 & 1673 & 42.13 & 13.35 & 1669 & 41.33 & 13.24 & 1658 \\
\hline
MMTLB + BCP         & 50.42 & 24.25 & 1916 & 48.09 & 23.83 & 1789 & 49.41 & 21.13 & 1790 & 49.62 & 21.45 & 2007 \\
\hline
TS-SLT(joint) + BCP & 49.02 & 24.66 & 1876 & 49.06 & 24.29 & 1833 & 50.28 & 21.56 & 1905 & 50.46 & 21.85 & 2042 \\
\hline
TS-SLT(ens.) + BCP  & 49.76 & 24.42 & 1839 & 49.39 & 24.63 & 1788 & 50.86 & 21.98 & 1899 & 51.23 & 22.24 & 2021 \\
\hline
CV-SLT + BCP        & \textbf{51.34} & \textbf{25.05} & 1808 & \textbf{50.46} & \textbf{24.88} & 1752 & \textbf{51.20} & \textbf{22.46} & 2175 & \textbf{51.65} & \textbf{22.78} & 2229 \\
\hline
\end{tabular}%
}
\end{table}

\subsection{Experimental Settings}
BCP is evaluated on PHOENIX-2014T~\cite{camgoz2018neural} (German) and CSL-Daily~\cite{zhou2021signbt} (Chinese), using the official train/dev/test splits. Both datasets are used for online evaluation of S2G, S2T, and S2G2T. Translation quality is measured by BLEU-4~\cite{papineni2002bleu} and ROUGE-L~\cite{lin2004rouge} for S2T and S2G2T, and WER for S2G; online latency is measured by Average Lagging (AL)~\cite{ma2019stacl} in milliseconds.

Following SimulSLT and CTL/CTL++, AL is still computed on the video timeline in milliseconds. The boundary-based advance unit only changes when the policy is allowed to advance; it does not change the metric definition or the unit used for comparison. This keeps AL directly comparable with the baselines.

The boundary adapter is a 2-layer MLP with hidden dimension 512 and GELU activation. Pseudo boundary labels are generated by TwoStream-SLR~\cite{chen2022twostream} via CTC forced alignment~\cite{graves2006ctc}. The adapter is trained with AdamW~\cite{loshchilov2019adamw} using a binary cross-entropy loss, learning rate $1\times10^{-4}$, batch size 16, for 30 epochs. The sweep is performed over $\tau_{thr} \in \{0.3, 0.5, 0.7\}$ and $K \in \{2, 3\}$, with default values $\tau_{thr}=0.3$ and $K=3$.

BCP is applied to MMTLB~\cite{chen2022mmtlb}, TwoStream-SLT~\cite{chen2022twostream} (joint and ensemble variants), and CV-SLT~\cite{zhao2024cvslt}, covering four offline methods on S2T and two on S2G/S2G2T. We compare BCP against SimulSLT~\cite{yin2021simulslt}, CTL, and CTL++~\cite{sun2024ctl} on S2T, and against SLRT~\cite{zuo2024online} on S2G and S2G2T. Following GMA~\cite{zhang2022gma}, we first summarize trainable parameters in Table~\ref{tab:params} and report the main fixed-latency S2T comparison in Table~\ref{tab:s2t}. We then present the full quality--latency trade-off in Figure~\ref{fig:tradeoff}, with S2G and S2G2T results discussed later in Table~\ref{tab:s2g_s2g2t}.

\subsection{Main Results}

\subsubsection{Sign-to-Text (S2T)}

\begin{figure}[t]
    \centering
    \includegraphics[width=0.6\linewidth]{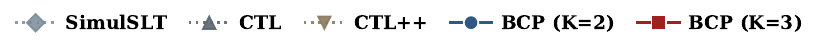}
    \begin{subfigure}[b]{0.48\linewidth}
        \centering
        \includegraphics[width=\linewidth]{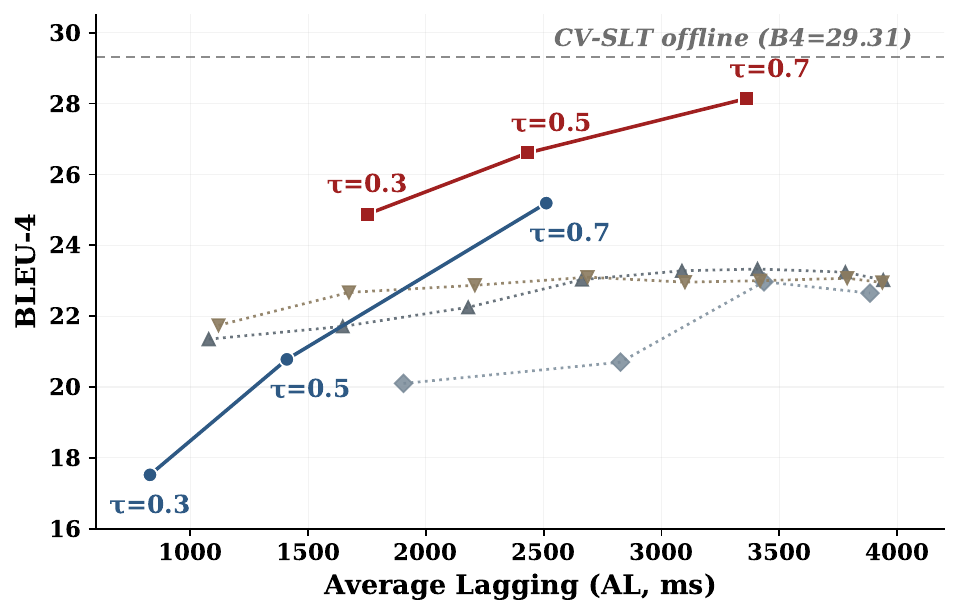}
        \caption{PHOENIX-2014T}
        \label{fig:tradeoff_phoenix}
    \end{subfigure}
    \hfill
    \begin{subfigure}[b]{0.48\linewidth}
        \centering
        \includegraphics[width=\linewidth]{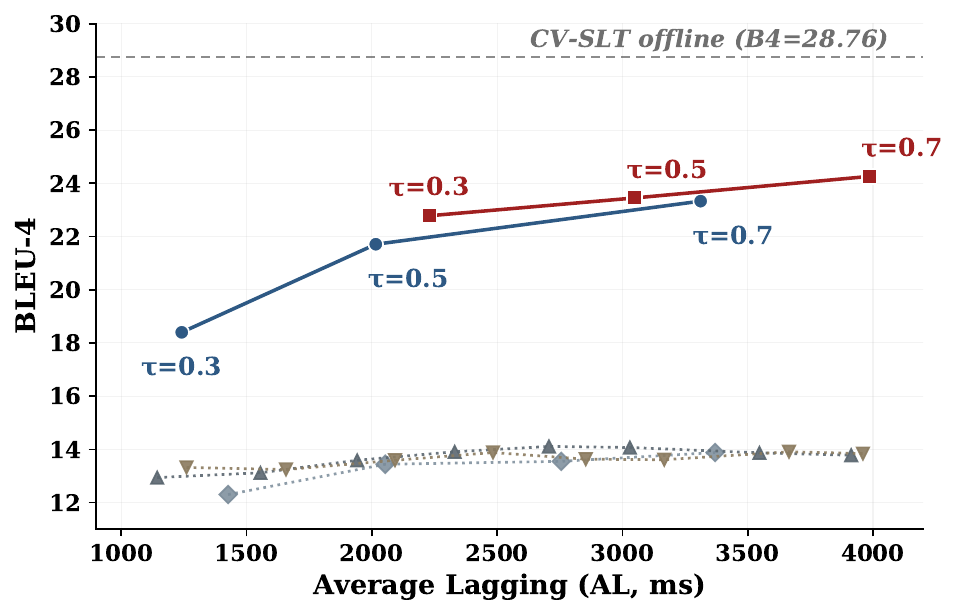}
        \caption{CSL-Daily}
        \label{fig:tradeoff_csl}
    \end{subfigure}
    \caption{Quality--latency trade-off of CV-SLT + BCP. The curves vary $\tau \in \{0.3, 0.5, 0.7\}$ with $K{=}2$ and $K{=}3$; dashed lines denote the offline upper bound of CV-SLT.}
    \label{fig:tradeoff}
\end{figure}

S2T is the primary task evaluated for BCP. Table~\ref{tab:params} summarizes trainable parameters; Table~\ref{tab:s2t} reports results under the default setting ($\tau{=}0.3, K{=}3$), and Figure~\ref{fig:tradeoff} shows the quality--latency trade-off.

\begin{table}[t]
\centering
\caption{Generalization of BCP to S2G and S2G2T. Best online results in bold.}
\label{tab:s2g_s2g2t}
\renewcommand{\arraystretch}{1.1}
\footnotesize
\begin{subtable}{\linewidth}
\centering
\setlength{\tabcolsep}{4pt}
\begin{tabular}{l|>{\centering\arraybackslash}p{1.4cm}|>{\centering\arraybackslash}p{1.4cm}|>{\centering\arraybackslash}p{1.4cm}|>{\centering\arraybackslash}p{1.4cm}}
\hline
\multirow{2}{*}{\textbf{Method (WER $\downarrow$)}} & \multicolumn{2}{c|}{\textbf{PHOENIX-2014T}} & \multicolumn{2}{c}{\textbf{CSL-Daily}} \\
\cline{2-3}\cline{4-5}
& Dev & Test & Dev & Test \\
\thickhline
\multicolumn{5}{l}{\textbf{Offline methods}} \\
\hline
MMTLB & 22.2 & 22.7 & 28.9 & 28.8 \\
\hline
TS-SLR & 17.6 & 19.2 & 25.5 & 25.3 \\
\hline
\multicolumn{5}{l}{\textbf{Online methods}} \\
\hline
SLRT & 22.2 & \textbf{22.1} & 30.2 & \textbf{29.3} \\
\hline
MMTLB + BCP & 25.2 & 25.5 & 31.6 & 31.9 \\
\hline
TS-SLT + BCP & \textbf{20.6} & 22.2 & \textbf{29.6} & 29.5 \\
\hline
\end{tabular}
\caption{Sign-to-Gloss (S2G) results.}
\label{tab:s2g}
\end{subtable}
\begin{subtable}{\linewidth}
\centering
\setlength{\tabcolsep}{3pt}
\begin{tabular}{l|c|c|c|c|c|c|c|c}
\hline
\multirow{3}{*}{\textbf{Method}} & \multicolumn{4}{c|}{\textbf{PHOENIX-2014T}} & \multicolumn{4}{c}{\textbf{CSL-Daily}} \\
\cline{2-5}\cline{6-9}
& \multicolumn{2}{c|}{Dev} & \multicolumn{2}{c|}{Test} & \multicolumn{2}{c|}{Dev} & \multicolumn{2}{c}{Test} \\
\cline{2-9}
& R & B4 & R & B4 & R & B4 & R & B4 \\
\thickhline
\multicolumn{9}{l}{\textbf{Offline methods}} \\
\hline
MMTLB & 49.98 & 24.46 & 49.78 & 25.02 & 51.13 & 21.67 & 51.64 & 21.52 \\
\hline
TS-SLT & 50.47 & 24.80 & 50.43 & 25.42 & 52.28 & 22.21 & 53.42 & 22.84 \\
\hline
\multicolumn{9}{l}{\textbf{Online methods}} \\
\hline
SLRT & 49.14 & \textbf{23.79} & 47.45 & \textbf{23.53} & 49.43 & \textbf{21.13} & 48.74 & 20.58 \\
\hline
MMTLB + BCP & 47.64 & 21.80 & 47.73 & 22.48 & 48.72 & 19.32 & 49.48 & 19.28 \\
\hline
TS-SLT + BCP & \textbf{49.39} & 23.75 & \textbf{48.55} & 23.03 & \textbf{50.56} & 20.84 & \textbf{50.04} & \textbf{20.69} \\
\hline
\end{tabular}
\caption{Sign-to-Gloss-to-Text (S2G2T) results.}
\label{tab:s2g2t}
\end{subtable}
\end{table}

\begin{table}[t]
\centering
\caption{Ablation study on PHOENIX-2014T (S2T, CV-SLT backbone). BP: boundary predictor; W: Wait-K with lag-$K$ constraint only and 8-frame video chunks as the advance unit; BCP: full method with LCP-based prefix voting. }
\label{tab:ablation}
\footnotesize
\renewcommand{\arraystretch}{1.1}
\setlength{\tabcolsep}{4pt}
\begin{tabular}{l|c|c|c|c|c|c}
\hline
\multirow{2}{*}{\textbf{Method}} & \multicolumn{3}{c|}{\textbf{Dev}} & \multicolumn{3}{c}{\textbf{Test}} \\
\cline{2-7}
& R & B4 & AL & R & B4 & AL \\
\thickhline
CV-SLT & 20.09 & 2.11 & $-$682 & 19.80 & 2.02 & $-$723 \\
\hline
CV-SLT + BP & 26.82 & 5.44 & $-$227 & 26.81 & 5.71 & $-$249 \\
\hline
CV-SLT + W & 34.06 & 9.94 & 132 & 34.16 & 10.45 & 128 \\
\hline
CV-SLT + BP + W & 52.44 & 25.87 & 2464 & 51.68 & 25.69 & 2375 \\
\hline
CV-SLT + BCP & 51.34 & 25.05 & 1808 & 50.46 & 24.88 & 1752 \\
\hline
\end{tabular}
\end{table}

BCP adds only 0.13M trainable parameters per backbone across S2T, S2G, and S2G2T, which is under 0.1\% of the 14.59M--436.12M frozen offline backbones and far below the 29--40M retrained by prior online methods. Despite this minimal footprint, BCP improves the S2T quality--latency trade-off in the evaluated settings. Under the default setting, all four BCP-adapted variants beat CTL++ on both Test sets. On PHOENIX-2014T, CV-SLT + BCP reaches 24.88 BLEU-4, exceeding CTL++ (22.67) by +2.21 at comparable latency (AL: 1752 vs.\ 1673~ms); on CSL-Daily, the margin widens to +9.54 (22.78 vs.\ 13.24). This larger gap likely reflects both the stronger frozen CV-SLT backbone on CSL-Daily and the fact that CTL++ was tuned mainly on German sign translation, whereas BCP's boundary-driven commit transfers more cleanly across languages. Even MMTLB + BCP, built on a weaker backbone, outperforms CTL++ on both Test sets, suggesting that the gain is not limited to the strongest backbone.

Figure~\ref{fig:tradeoff} further shows that BCP generally offers a better quality--latency trade-off than the online baselines. On PHOENIX-2014T, the baselines saturate beyond AL$\approx$2700~ms (best BLEU-4: 23.33), whereas BCP reaches 28.14 BLEU-4 at AL$\approx$3360~ms. On CSL-Daily, its best setting reaches 24.26 BLEU-4, staying above the baselines. At low latency ($K{=}2, \tau{=}0.3/0.5$), BCP is close to the baselines because prefix voting has too few candidates; this matches the no-slower-than-Wait-K design.

\subsubsection{Generalization to S2G and S2G2T}
To verify cross-task generality, BCP is evaluated on S2G and S2G2T using the same boundary predictor trained for S2T, without any task-specific retraining or additional parameters. Table~\ref{tab:s2g_s2g2t} compares BCP with the online baseline SLRT.

BCP remains competitive with SLRT on S2G. On S2G (Table~\ref{tab:s2g}), BCP(TS-SLR) reaches 20.6/22.2 WER on PHOENIX-2014T Dev/Test and 29.6/29.5 on CSL-Daily, comparable to SLRT (22.2/22.1 and 30.2/29.3). The shared boundary predictor transfers to S2G without task-specific tuning.

BCP improves ROUGE-L while remaining comparable in BLEU-4 on S2G2T. On S2G2T (Table~\ref{tab:s2g2t}), BCP(TS-SLT) improves ROUGE-L over SLRT across all four evaluation splits (e.g., 50.04 vs.\ 48.74 on CSL-Daily Test) and achieves comparable BLEU-4. The performance gap between BCP(MMTLB) and BCP(TS-SLR/TS-SLT) reflects backbone-dependent scaling; stronger offline backbones yield better online output. These results suggest that BCP can transfer from S2T to both auxiliary tasks with a single shared predictor.

\subsection{Component Analysis}

Table~\ref{tab:ablation} analyzes BCP on PHOENIX-2014T (S2T, CV-SLT backbone) by progressively adding components from the offline backbone toward the full method.

Without online control, CV-SLT on partial video input produces near-random output (Test B4 = 2.02) with strongly negative AL ($-$723~ms), because the offline decoder overproduces tokens from limited input. Adding the boundary predictor (BP) alone provides timing signals but no commit policy: quality improves to B4 = 5.71 yet AL remains negative ($-$249~ms).

Applying Wait-K alone (W), with incoming 8-frame video chunks as the advance unit, yields positive AL (128~ms) and B4 = 10.45; chunk-based pacing aligns the target with the source temporally but limits translation quality. Combining BP and Wait-K turns boundaries into well-defined advance units, raising B4 substantially to 25.69 at AL = 2375~ms. This confirms that BP supplies the semantic timing signal that Wait-K requires to operate effectively on continuous video.

Adding LCP voting to form the full BCP reduces AL from 2375~ms to 1752~ms ($-$623~ms, a 26\% reduction) with only a minor B4 change ($-$0.81), showing that it lowers effective latency without sacrificing quality. Together, BP decides when, Wait-K provides the fallback progress schedule, and LCP voting accelerates commits when a longer stable prefix is available.

We also checked sensitivity to the frozen CTC label source. Replacing TS-SLR with MMTLB, while keeping the same PHOENIX dev setup, changes boundary F1 from 0.7079 to 0.7324, B4 from 25.05 to 25.31, and AL from 5.65 to 5.72 steps. The small downstream change suggests that BCP is reasonably robust to the auxiliary model choice because the alignment mostly captures timing on the shared gloss sequence.

\section{Conclusion}
This paper presents BCP, a lightweight framework that converts pretrained offline sign language models to online use. BCP uses a CTC-supervised boundary adapter to decide when to commit and a Boundary-Controlled Wait-K policy with LCP voting to decide what to commit, yielding burst commits while preserving base Wait-K progress. Experiments on PHOENIX-2014T and CSL-Daily show a strong S2T quality--latency trade-off and competitive S2G/S2G2T results with under 0.1\% trainable parameters per method. Since all backbones remain frozen, BCP can extend to future offline models without task-specific data construction; future work will study longer streams and alternative boundary supervision.

\begin{credits}
\subsubsection{\ackname}
This work is partially supported by NSFC, China (No. 62276196), and Guangxi Science and Technology Major Program (No. AA24206067).
\end{credits}

%
% ---- Bibliography ----
%
% BibTeX users should specify bibliography style 'splncs04'.
\bibliographystyle{splncs04}
\bibliography{references}
% References will then be sorted and formatted in the correct style.
%

\end{document}